\documentclass[conference]{IEEEtran}
\IEEEoverridecommandlockouts

\usepackage{amsmath} 
\usepackage{amsmath,amsfonts}
\usepackage{algpseudocode}
\usepackage{algorithm}
\usepackage{array}
\usepackage{textcomp}
\usepackage{stfloats}
\usepackage{url}
\usepackage{verbatim}
\usepackage{graphicx}
\usepackage{cite}
\usepackage{caption}
\usepackage{multirow}
\usepackage{colortbl} 
\usepackage{xcolor}
\usepackage{subcaption}
\usepackage{arydshln}
\usepackage{placeins}
\usepackage{tabularray}
\UseTblrLibrary{booktabs}
\usepackage{hyperref}
\usepackage{amssymb}
\usepackage{tikz}
\usetikzlibrary{arrows.meta, shapes.geometric, positioning}
\usepackage{pgfplots}
\usepackage[letterpaper, top=54pt, bottom=73pt, left=45pt, right=45pt, nohead, nofoot]{geometry}
\usepgfplotslibrary{groupplots}
\usetikzlibrary{calc}
\pgfplotsset{compat=1.18} 
\usepackage{tcolorbox}
\tcbuselibrary{skins, breakable}

\title{\LARGE \bf
Multimodal-Language-Model–Driven Interaction and Companionship for Service Robots in Elderly-Care Facilities
}

\author{Ching-Chieh Liu, Cong-Thanh Vu and Yen-Chen Liu
\thanks{This work was supported in part by the National Science and Technology Council (NSTC), Taiwan, under Grants NSTC 114-2628-E-006-010 and NSTC 114-2218-E-006-021, and in part by the Higher Education Sprout Project of the Ministry of Education to the Headquarters of University Advancement at National Cheng Kung University (NCKU).}
\thanks{All authors are with the Department of Mechanical Engineering, National Cheng Kung University (NCKU), Tainan 70101, Taiwan. Email: {\href{mailto:n16130289@gs.ncku.edu.tw}{\texttt{n16130289@gs.ncku.edu.tw}}, \href{mailto:vuthanh.cdt@gmail.com} {\texttt{vuthanh.cdt@gmail.com}}}, \href{mailto:yliu@mail.ncku.edu.tw}{\texttt{yliu@mail.ncku.edu.tw.}}}}

\begin{document}

\maketitle
\thispagestyle{empty}
\pagestyle{empty}

\begin{abstract}
Service robots are increasingly deployed in elderly-care facilities to alleviate caregiver workload and enhance the quality of daily care. However, most existing studies focus on isolated service functions and lack integrated capabilities for continuous companionship, natural interaction, and safety monitoring. In this paper, we present an intelligent companion robot system that unifies active visual human-following, real-time LLM-driven speech interaction for intent understanding and task execution, and VLM-based safety monitoring for fall detection and abnormal posture assessment. The perception layer ensures robust human tracking and uses an active gimbal to maintain the user in view during occlusions or abrupt movements. At the interaction layer, a Large Language Model interprets spoken requests and maps them to robot actions, enabling escorting and semantic navigation. Simultaneously, a VLM-based safety agent continuously analyzes visual observations to detect fall-related or abnormal postures and triggers emergency responses when necessary. Experimental results demonstrate the system’s ability to reliably follow and interact with humans, while effectively detecting potential falls to ensure user safety.
\end{abstract}

\section{Introduction}
\begin{figure}
\centering
\includegraphics[scale=0.37,page=1]{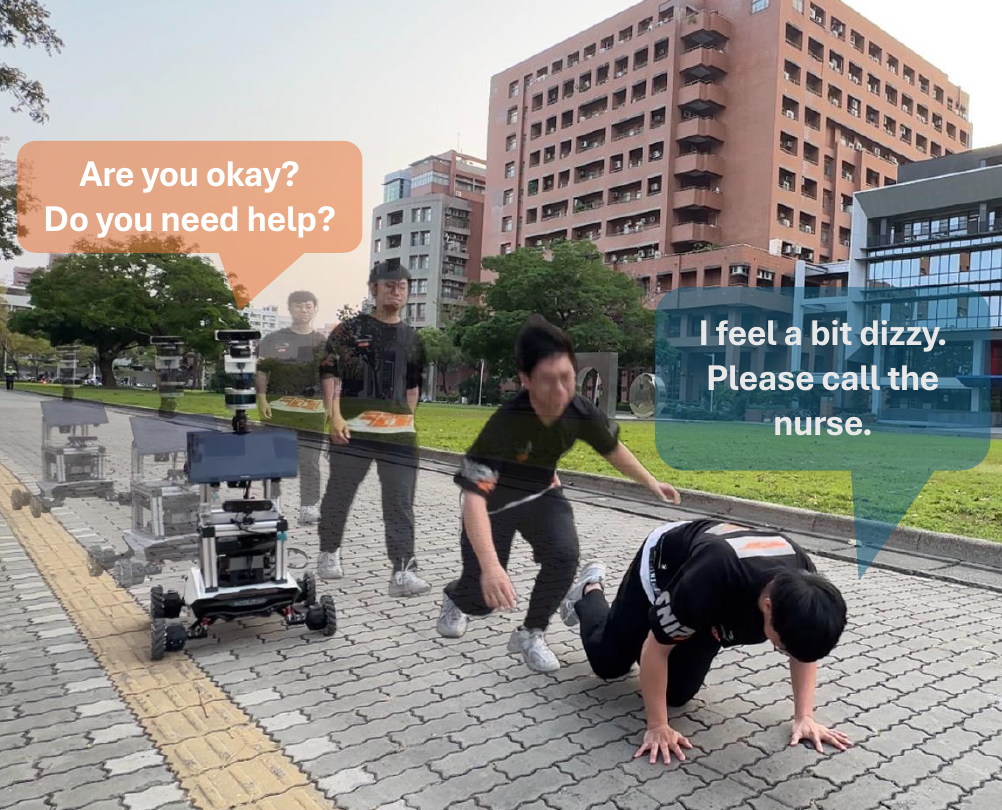}
\caption{The proposed companion robot integrating visual tracking, LLM interaction, and VLM fall detection.}
\label{fig:HRI.pdf}
\end{figure}

Mobile robots are being deployed widely across numerous application domains~\cite{10.1007/978-981-16-2094-2_49}, particularly in service-oriented environments. Service robots are increasingly utilized in elderly-care facilities to alleviate caregiver workloads by providing physical accompaniment and continuous safety monitoring~\cite{Li2023}. Specifically, these systems are seeing widespread adoption to both reduce staff burden and enhance the overall quality of resident care~\cite{Chen2022, Islam2019}. In such settings, residents often require physical accompaniment, guided navigation between functional areas, and rapid response during health-related events. Fig.~\ref{fig:HRI.pdf} illustrates a representative companionship scenario in which a user initiates a request through natural speech, and the robot responds by providing an escort while maintaining persistent visual tracking, as well as delivering emergency assistance such as fall detection and alerting. Given ongoing staffing shortages and heavy caregiving demands, elderly-care facilities necessitate robust robotic solutions capable of delivering simultaneous mobility assistance and continuous safety supervision.

Human-following capabilities form the foundation of this mission, enabling mobile companionship and interactive assistance~\cite{Li2023, Islam2019}. The field has progressed from simple follow-behind behaviors~\cite{VanToan2023} to more advanced side-by-side accompaniment strategies~\cite{Xue2022}. Recent developments further enable robots to autonomously determine socially appropriate positioning through environment-aware reasoning~\cite{11093565, 11246444} and perform socially compliant maneuvers that improve user comfort~\cite{Peng2024}.

Despite these advances, human-following alone is insufficient. Since elderly residents often express needs indirectly, robots must move beyond rigid interfaces and use natural dialogue to clarify intentions~\cite{Marge2022}. Additionally, emergency detection like fall risk must be integrated into the core control loop rather than treated as a decoupled module. Therefore, tightly coupling natural spoken interaction with real-time state assessment is essential to resolve ambiguities and ensure timely assistance.

\begin{figure*}[ht!]
    \centering
    \includegraphics[width=0.93\linewidth]{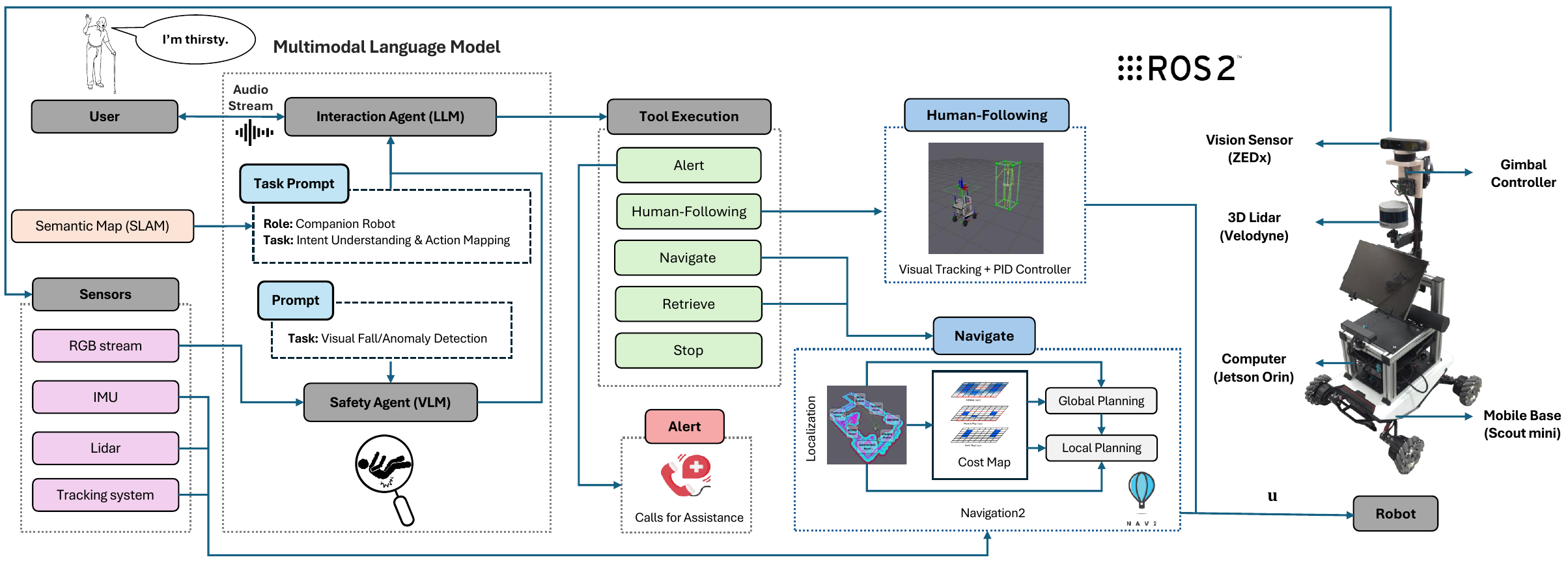}
    \caption{System architecture of the proposed companion robot: an LLM agent generates constrained actions, while a parallel VLM agent handles fall detection and emergency overrides.}
    \label{fig:system.pdf}
\end{figure*}
Recent progress in Large Language Models (LLMs) and Vision–Language Models (VLMs) has catalyzed language-driven task specification and tool-oriented execution in robotics. Prior work has demonstrated the translation of natural language into executable robot programs~\cite{Venkatesh2021}, the use of LLMs as policies for embodied control~\cite{Liang2023}, and situated task planning through prompt engineering~\cite{Singh2023}. Beyond high-level planning, LLMs are increasingly used for interaction adaptation~\cite{Su2024} and for mapping user instructions to feasible motion plans~\cite{Lin2023}.

While individual components such as tracking, state estimation, and LLM-based reasoning have matured significantly, a critical gap persists because safety monitoring is still largely treated as an external add-on. There is therefore a pressing need for integrated architectures in which vision-based agents operate continuously, executing interactive following behaviors while simultaneously providing preemptive real-time safety monitoring and emergency assistance.

To address these challenges, we propose a companion robot designed specifically for elderly-care applications that integrates three tightly coupled components: (i) active visual tracking, (ii) LLM-based dialogue interaction, and (iii) VLM-based safety monitoring. The main contributions of this work are summarized as follows:
\begin{itemize}
\item We develop a visual tracking system integrated with a gimbal mechanism to maintain the user within the camera’s field of view regardless of movement or orientation. This perception module is further coupled with a tracking controller to ensure stable and continuous human-following performance.
\item We design an interaction layer centered on an LLM that performs conversational intent understanding and maps user requests to constrained robot functions and semantic navigation goals. This enables semantic navigation and caregiving assistance beyond traditional geometry-based distance regulation.
\item We propose a decoupled safety agent that employs a VLM to assess fall-related and abnormal postural states from visual observations. Upon detection, the agent immediately overrides ongoing behaviors and triggers an emergency response, enabling timely safety intervention in elderly-care facilities.
\end{itemize}
\section{System Design}\label{sec:system_design}
This section presents the experimental robotic platform, including its hardware and software components, with the overall system architecture illustrated in Fig.~\ref{fig:system.pdf}.
\subsection{Hardware Architecture}
The robotic system is built upon the AgileX SCOUT MINI mobile platform ($612 \times 580 \times 245$ mm), featuring a mecanum-wheel configuration that enables holonomic maneuverability. At the core of the architecture is an NVIDIA Jetson Orin embedded module (2048 CUDA cores, 275 TOPS), which serves as the high-performance computing hub. This unit enables the real-time integration of computationally intensive tasks, including deep-learning-based visual perception, SLAM-based localization, and low-latency communication with cloud-hosted LLMs for natural language processing.

Environmental perception is achieved through the fusion of a Velodyne VLP-16 3D LiDAR for localization and obstacle-aware navigation, together with a ZED X stereo camera interfaced via GMSL2. The camera is integrated with a single-motor active gimbal dedicated to user-centric monitoring and human tracking. This gimbal-mounted camera also utilizes a VLM to continuously analyze the user’s posture and activities, enabling the detection of safety-critical events such as falls in elderly care scenarios. A PID-based tracking controller dynamically adjusts the gimbal orientation to keep the user centered in the frame, ensuring reliable VLM inference even during rapid robot maneuvers. Finally, a DJI Mic wireless system captures speech while mitigating robot ego-noise, paired with an onboard loudspeaker to facilitate natural bidirectional interaction.
\subsection{Software Architecture}
The software architecture is implemented using the ROS 2 framework and is organized into five core functional components: (1) skeleton detection and 3D target localization, (2) following control, (3) localization and mapping, (4) semantic navigation, and (5) multimodal LLM/VLM interaction and safety monitoring. Initially, the ZED SDK~\cite{zed_sdk} performs real-time human detection and skeleton tracking from the gimbal-mounted camera stream. These visual cues are fused with depth data to estimate the target’s relative 3D coordinates within the robot’s coordinate frame, providing the necessary input for the escorting control law. Simultaneously, localization is performed using Fast LIO 2~\cite{9697912}, ensuring consistent state estimation within the global map frame.

For autonomous navigation, the system leverages the ROS 2 Navigation2 (Nav2) \cite{macenski2020marathon2} stack to perform global and local path planning, obstacle avoidance, and motion execution based on goal poses and velocity commands. For human–robot interaction, the system employs the GPT-4o Realtime API as its core speech engine. A dedicated Python-based ROS 2 node manages bidirectional low-latency audio streaming while maintaining conversational context to ensure responsive and natural dialogue. Running in parallel, a VLM-based safety agent continuously interprets visual observations. This agent has the authority to preempt ongoing behaviors and trigger emergency escalation protocols upon detecting safety-critical anomalies, such as user falls.
\begin{figure}[t]
\centering
\includegraphics[scale=0.42,page=1]{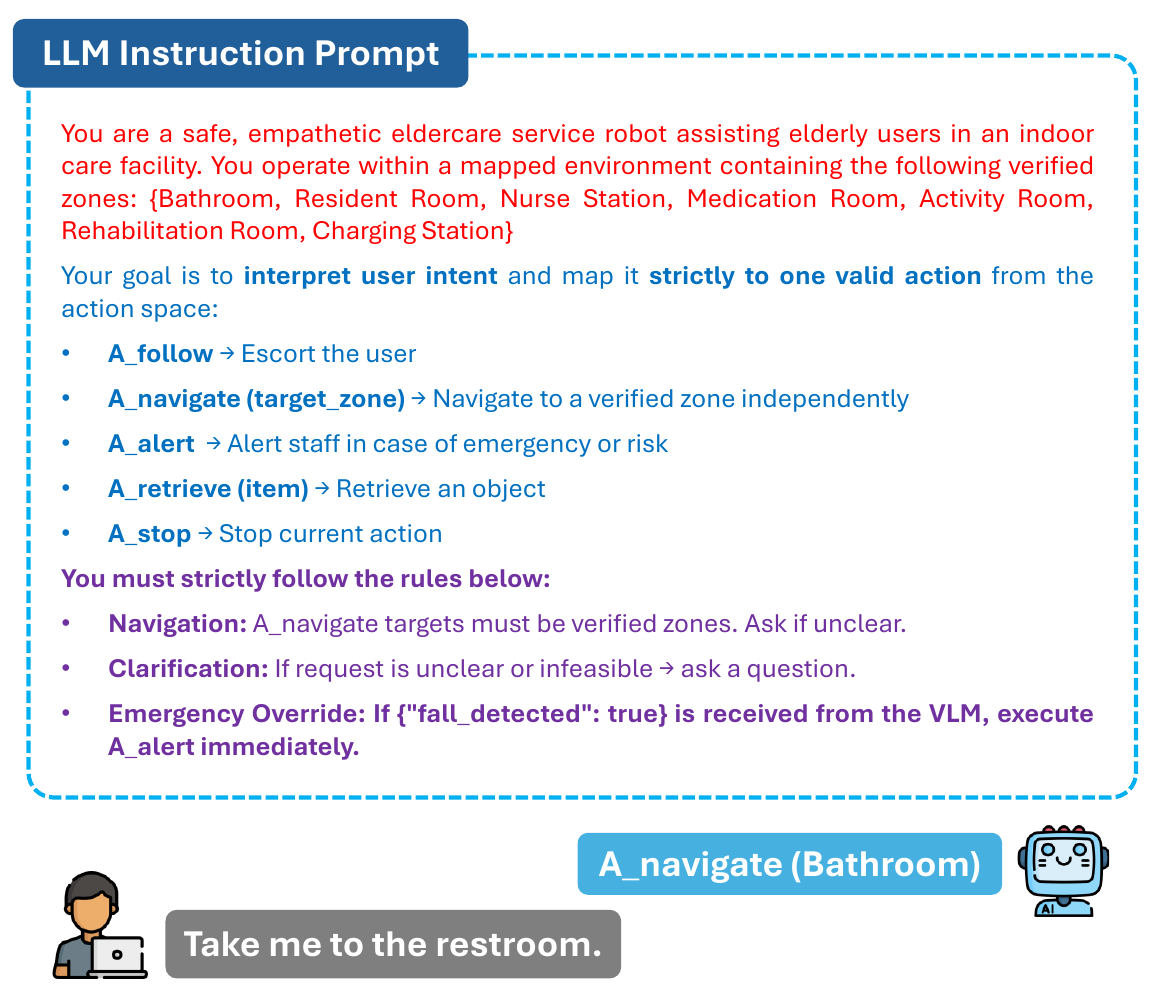}
\caption{Structured system prompt for interaction.}
\label{fig:llm_prompt}
\end{figure}
\section{Methodology}\label{sec:methodology}
In this section, we present the proposed methodology, which comprises three core integrated layers: (i)Visual Perception and Human-Following Control; (ii) an LLM-driven interaction layer for dialogue grounding and autonomous tool invocation; and (iii) a VLM-based safety layer for fall-related state assessment and emergency escalation.

\subsection{Visual Perception and Human-Following Control}
\label{subsec:perception_and_control}

Our human-following module utilizes a gimbal-mounted ZED~X stereo camera and the ZED SDK to directly extract the user's 3D spatial coordinates. To minimize target loss during rapid movements, the gimbal actively stabilizes the camera. Using the target's horizontal pixel location $u$ and the image center $c_x$, the controller calculates the horizontal error $e_u = u - c_x$. A PID control law then updates the gimbal yaw to drive $e_u$ toward zero, ensuring continuous visual contact.

Once acquired, the human's coordinates are obtained in the robot's local frame $\mathcal{R}$. Let the robot and human states in the global frame $\mathcal{G}$ be $\mathbf{q}^r = [x^r, y^r, \theta^r]^{\mathrm{T}} \in \mathbb{R}^3$ and $\mathbf{q}^h = [x^h, y^h, \theta^h]^{\mathrm{T}} \in \mathbb{R}^3$, respectively. The human's state relative to $\mathcal{R}$ is denoted by $\mathbf{q}_h^r = [x_h^r, y_h^r, \theta_h^r]^{\mathrm{T}} \in \mathbb{R}^3$. The coordinate transformation from $\mathcal{R}$ to $\mathcal{G}$ is:
\begin{equation}
    \mathbf{q}^h = {}_{\mathcal{R}}^{\mathcal{G}}\mathbf{T} \mathbf{q}_h^r,
    \label{eq:coordinate_transform}
\end{equation}
where
\begin{equation*}
    {}_{\mathcal{R}}^{\mathcal{G}}\mathbf{T} = 
    \begin{bmatrix} 
        \mathbf{J}(\theta^r) & \mathbf{q}^r \\ 
        0 & 1 
    \end{bmatrix}, \quad 
    \mathbf{J}(\theta^r) = 
    \begin{bmatrix} 
        \cos\theta^r & -\sin\theta^r & 0 \\ 
        \sin\theta^r & \cos\theta^r & 0 \\ 
        0 & 0 & 1 
    \end{bmatrix},
\end{equation*}
${}_{\mathcal{R}}^{\mathcal{G}}\mathbf{T}$ is the transformation matrix from $\mathcal{R}$ to $\mathcal{G}$, and $\mathbf{J}(\theta^r)$ is the rotation transformation matrix.

To support language-driven companionship, the robot uses a mode-gated architecture: the Nav2 stack~\cite{macenski2020marathon2} handles global point-to-point navigation, while a relative-tracking controller~\cite{11246444} manages active escorting. The tracking controller adjusts the relative position $(x_h^r, y_h^r)$ to maintain a desired tracking pose $\mathbf{q}^d = [x^d, y^d, \theta^d]^{\mathrm{T}} \in \mathbb{R}^3$, ensuring $\lim_{t \to \infty} |\mathbf{q}^r - \mathbf{q}^d| = 0$.
\subsection{LLM-Driven Dialogue Interaction and Tool Invocation}\label{subsec:llm}
The interaction layer employs an LLM to translate unstructured spoken requests into constrained, safety-verified robot actions. As depicted in Fig.~\ref{fig:system.pdf}. This is formulated as a context-grounded decision process with the system state defined as:
\begin{equation}
\mathcal{S}(t)=\{I(t), C(t), M\},
\end{equation}
where $I(t)$ denotes the user utterance, $C(t)$ encodes the operating mode, and $M$ is the semantic map containing verified functional zones. Given $\mathcal{S}(t)$ and a structured system prompt as described in Fig.~\ref{fig:llm_prompt}, the LLM generates a discrete high-level action expressed as
\begin{equation}
a = f_{\mathrm{LLM}}(\mathcal{S}(t), \mathrm{Prompt}_{\mathrm{sys}}),
\end{equation}
where $a$ represents a specific executable command selected from the validated action set $\mathcal{A}$. The action set is strictly bounded to ensure predictable execution and to prevent free-form actuation, with each action deterministically mapped to ROS~2 tool function calls, as detailed in Table~\ref{tab:action_mapping}.

\begin{figure}[t]
    \centering
    \includegraphics[width=0.75\linewidth]{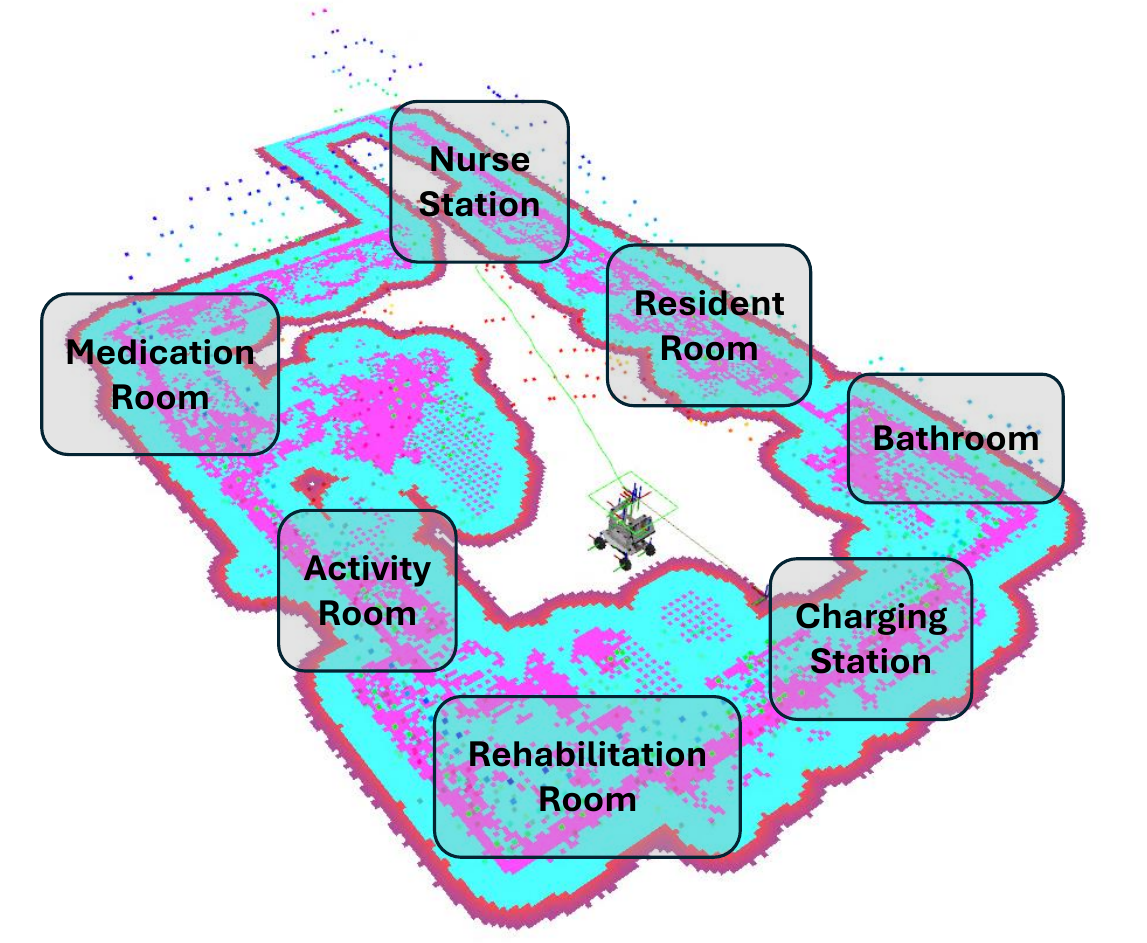}
    \caption{Semantic SLAM map linking annotated functional zones to unique global poses for deterministic navigation.}
    \label{fig:nav_map}
\end{figure}

\begin{table}[ht]
\centering
\caption{Dialogue-to-action mapping for the LLM interaction agent.}
\renewcommand{\arraystretch}{1.9}
\begin{tabular}{||m{1.5cm}||m{5.7cm}||}
\hline
\textbf{Action} & \textbf{Functional Description} \\ 
\hline\hline
\texttt{A\_follow} & Activates the active human-following mode. \\ 
\hline
\texttt{A\_navigate} & Navigates to a specified destination. \\ 
\hline
\texttt{A\_alert} & Detects a user emergency and calls for assistance. \\ 
\hline
\texttt{A\_retrieve} & Navigates to retrieve an item for the user. \\
\hline
\texttt{A\_stop} & Immediately stops. \\ 
\hline
\end{tabular}
\label{tab:action_mapping}
\end{table}

To ensure safety and prevent unpredictable outputs in caregiving environments, the LLM is strictly confined to a predefined tool library with explicit argument schemas. Rather than allowing open-ended generation, the system incorporates zone labels from the SLAM map $M$, as illustrated in Fig.~\ref{fig:nav_map}, as hard environmental constraints. Consequently, a \texttt{A\_navigate} command is accepted only if it targets a recognized region; once verified, the corresponding goal pose is sent to Nav2 for execution. Furthermore, the system continuously evaluates every generated tool call at runtime. Invalid or contextually inappropriate actions, such as attempting to navigate during an emergency state, are immediately blocked and redirected into a clarification dialogue with the user. By restricting the LLM to this architecture, we minimize execution ambiguity and ensure that the robot’s decisions remain tightly aligned with its physical state and established care workflows.

\subsection{VLM-Based Safety Monitoring and Emergency Alert}
\label{subsec:vlm}
\begin{figure}[ht]
\centering
\includegraphics[scale=0.34,page=1]{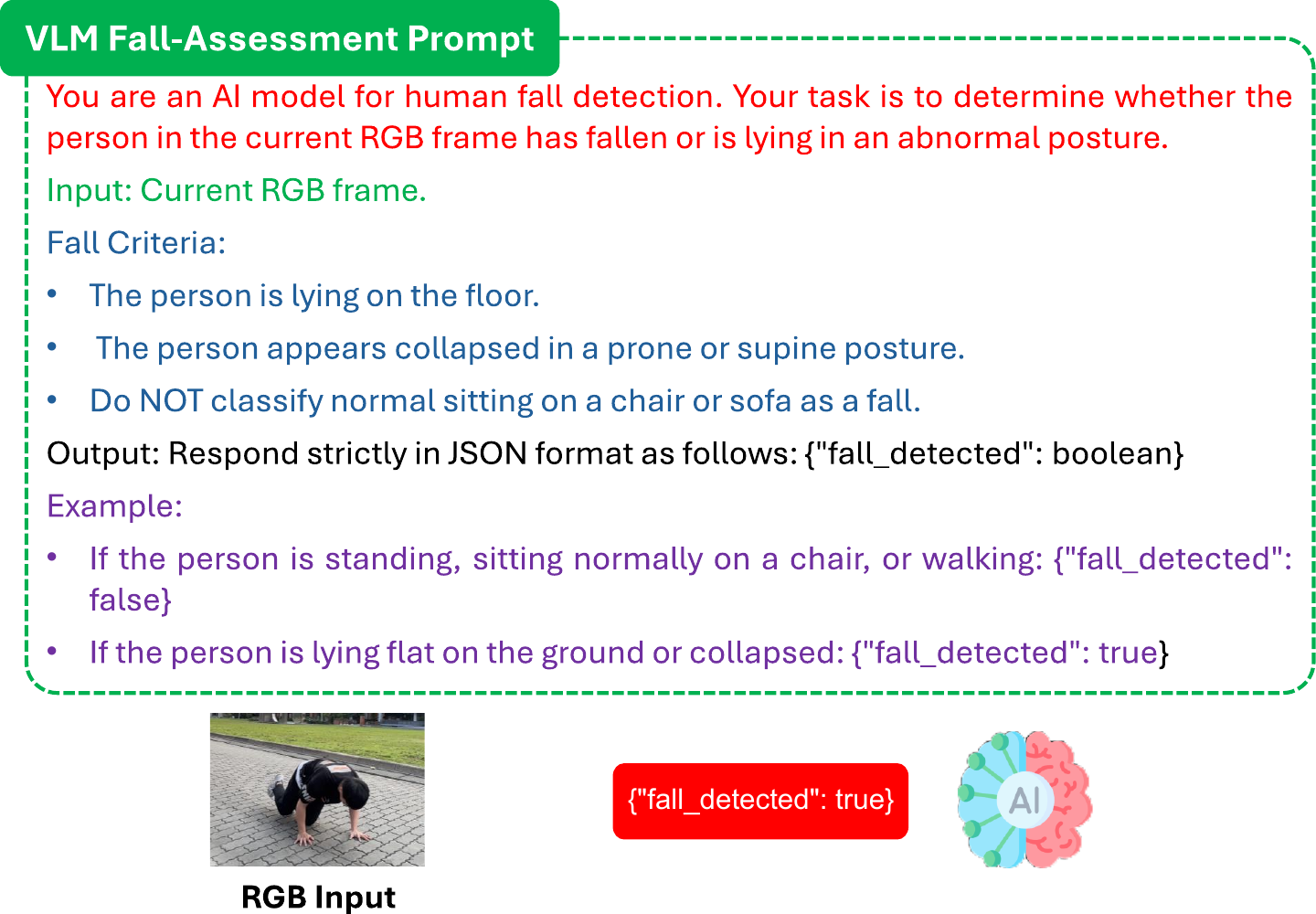}
\caption{Structured VLM prompt for fall assessment.}
\label{fig:vlm_fall_prompt}
\end{figure}

To monitor fall-related incidents and abnormal postural states of the user, a dedicated safety agent runs in parallel with the interaction agent using the proposed VLM. The agent captures RGB observations $I_{\mathrm{rgb}}(t)$ and queries the VLM with a structured fall-assessment prompt, as shown in Fig.~\ref{fig:vlm_fall_prompt}. The emergency problem can be formally expressed as follows:
\begin{equation}
F(t) = f_{\mathrm{VLM}}\!\left(I_{\mathrm{rgb}}(t),\,\mathrm{Prompt}_{\mathrm{fall}}\right),
\end{equation}
where $F(t)\in\{0,1\}$ indicates whether an abnormal state is detected. If $F(t)=1$, the safety agent asserts an emergency state, interrupts any ongoing escorting or navigation behavior, and publishes the anomaly event via ROS to the LLM interaction layer. The interaction layer then issues the $\texttt{A\_alert}$ action to notify caregivers and escalate the situation. This decoupled design ensures persistent safety monitoring independent of dialogue execution, while providing a reliable hard-override pathway for timely intervention.

\section{Results and Discussion}\label{sec:results}
In this section, we present an evaluation of the proposed method using the robotic platform described in Section~\ref{sec:system_design}. The experiments cover four key aspects of system performance: the tracking capability of the human-following controller, the execution of semantic interaction tasks in an elderly-care context, a subjective user study assessing interaction quality, and the effectiveness of the VLM-based fall detection module.
\subsection{Tracking Performance}
\label{subsec:exp_following}
To evaluate the robustness of the human-following controller and the active gimbal system, tracking experiments were conducted on a $100\,\mathrm{m}$ everyday course comprising straight segments and curves over brick and asphalt surfaces. Three individuals participated in this study, each walking the entire distance at three distinct speeds. The three speed levels were defined as follows: low ($0.7\,\mathrm{m/s}$), normal ($1.2\,\mathrm{m/s}$), and high ($1.5\,\mathrm{m/s}$) walking speeds. 

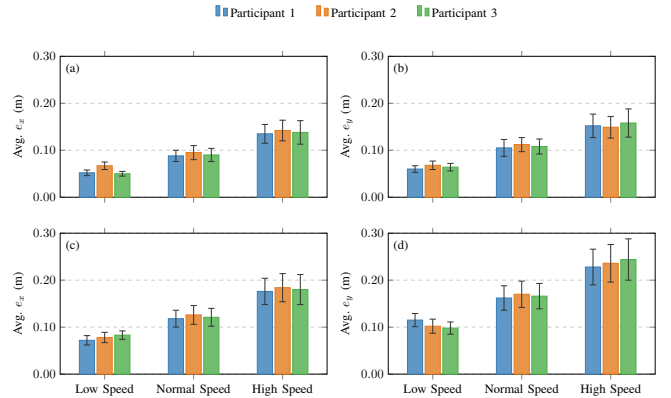
\begin{figure}[ht]
\centering
\resizebox{0.95\columnwidth}{!}{%
\begin{tikzpicture}

\definecolor{P1Blue}{RGB}{55, 126, 184}
\definecolor{P2Orange}{RGB}{228, 122, 28}
\definecolor{P3Green}{RGB}{77, 175, 74}

\begin{groupplot}[
    group style={
        group size=2 by 2,
        horizontal sep=1.7cm,
        vertical sep=1.0cm,
    },
    width=9cm,
    height=5.5cm,
    every axis/.append style={
        ybar=2pt,
        bar width=12pt,
        scaled y ticks=false,
        yticklabel style={
            /pgf/number format/fixed,
            /pgf/number format/precision=2,
            /pgf/number format/fixed zerofill
        }
    },
    enlarge x limits=0.25,
    ymin=0, ymax=0.30,
    symbolic x coords={Low Speed, Normal Speed, High Speed},
    xtick=data,
    ymajorgrids=true,
    grid style={dashed, gray!40},
    tick label style={font=\normalsize},
    label style={font=\normalsize},
    error bars/y dir=both,
    error bars/y explicit,
    error bars/error bar style={line width=0.8pt, black!80},
    error bars/error mark options={rotate=90, mark size=2.5pt, line width=0.8pt},
]

\nextgroupplot[
    ylabel={Avg. $e_x$ (m)},
    xticklabels={,,}, 
    legend to name=TrackingLegend,
    legend columns=-1,
    legend style={
        draw=none,
        fill=none,
        font=\normalsize,
        /tikz/every even column/.append style={column sep=0.5cm}
    },
]

\addplot[fill=P1Blue!80, draw=P1Blue!100, line width=0.8pt] coordinates {
    (Low Speed, 0.052) +- (0, 0.006)
    (Normal Speed, 0.088) +- (0, 0.012)
    (High Speed, 0.135) +- (0, 0.020)
}; \addlegendentry{Participant 1}

\addplot[fill=P2Orange!80, draw=P2Orange!100, line width=0.8pt] coordinates {
    (Low Speed, 0.067) +- (0, 0.008)
    (Normal Speed, 0.095) +- (0, 0.015)
    (High Speed, 0.142) +- (0, 0.022)
}; \addlegendentry{Participant 2}

\addplot[fill=P3Green!80, draw=P3Green!100, line width=0.8pt] coordinates {
    (Low Speed, 0.050) +- (0, 0.005)
    (Normal Speed, 0.090) +- (0, 0.014)
    (High Speed, 0.138) +- (0, 0.025)
}; \addlegendentry{Participant 3}

\nextgroupplot[
    ylabel={Avg. $e_y$ (m)},
    xticklabels={,,}, 
]

\addplot[fill=P1Blue!80, draw=P1Blue!100, line width=0.8pt] coordinates {
    (Low Speed, 0.060) +- (0, 0.007)
    (Normal Speed, 0.105) +- (0, 0.018)
    (High Speed, 0.152) +- (0, 0.025)
};

\addplot[fill=P2Orange!80, draw=P2Orange!100, line width=0.8pt] coordinates {
    (Low Speed, 0.068) +- (0, 0.009)
    (Normal Speed, 0.112) +- (0, 0.015)
    (High Speed, 0.149) +- (0, 0.023)
};

\addplot[fill=P3Green!80, draw=P3Green!100, line width=0.8pt] coordinates {
    (Low Speed, 0.064) +- (0, 0.008)
    (Normal Speed, 0.108) +- (0, 0.016)
    (High Speed, 0.158) +- (0, 0.030)
};

\nextgroupplot[
    ylabel={Avg. $e_x$ (m)},
]

\addplot[fill=P1Blue!80, draw=P1Blue!100, line width=0.8pt] coordinates {
    (Low Speed, 0.072) +- (0, 0.010)
    (Normal Speed, 0.118) +- (0, 0.018)
    (High Speed, 0.176) +- (0, 0.028)
};

\addplot[fill=P2Orange!80, draw=P2Orange!100, line width=0.8pt] coordinates {
    (Low Speed, 0.078) +- (0, 0.011)
    (Normal Speed, 0.126) +- (0, 0.020)
    (High Speed, 0.184) +- (0, 0.030)
};

\addplot[fill=P3Green!80, draw=P3Green!100, line width=0.8pt] coordinates {
    (Low Speed, 0.083) +- (0, 0.009)
    (Normal Speed, 0.121) +- (0, 0.019)
    (High Speed, 0.180) +- (0, 0.032)
};

\nextgroupplot[
    ylabel={Avg. $e_y$ (m)},
]

\addplot[fill=P1Blue!80, draw=P1Blue!100, line width=0.8pt] coordinates {
    (Low Speed, 0.115) +- (0, 0.014)
    (Normal Speed, 0.162) +- (0, 0.026)
    (High Speed, 0.228) +- (0, 0.038)
};

\addplot[fill=P2Orange!80, draw=P2Orange!100, line width=0.8pt] coordinates {
    (Low Speed, 0.102) +- (0, 0.015)
    (Normal Speed, 0.170) +- (0, 0.028)
    (High Speed, 0.236) +- (0, 0.040)
};

\addplot[fill=P3Green!80, draw=P3Green!100, line width=0.8pt] coordinates {
    (Low Speed, 0.098) +- (0, 0.013)
    (Normal Speed, 0.166) +- (0, 0.027)
    (High Speed, 0.244) +- (0, 0.044)
};

\end{groupplot}

\node at ($(group c1r1.north)!0.5!(group c2r1.north) + (0, 1.2cm)$) {\ref{TrackingLegend}};

\node[font=\normalsize] at ($(group c1r1.north west)+(0.35cm,-0.35cm)$) {(a)};
\node[font=\normalsize] at ($(group c2r1.north west)+(0.35cm,-0.35cm)$) {(b)};
\node[font=\normalsize] at ($(group c1r2.north west)+(0.35cm,-0.35cm)$) {(c)};
\node[font=\normalsize] at ($(group c2r2.north west)+(0.35cm,-0.35cm)$) {(d)};

\end{tikzpicture}%
}
\caption{Human-following performance at various velocities: (a, b) with and (c, d) without gimbal stabilization.}
\label{fig:tracking_error_bar}
\end{figure}

The evaluation reveals a clear correlation between target velocity and tracking errors, as shown in Fig.~\ref{fig:tracking_error_bar}, tracking errors increase with velocity; however, gimbal stabilization significantly mitigates these deviations. Without the gimbal (Fig.~\ref{fig:tracking_error_bar}c, d), the lateral error ($e_y$) surges at high speeds due to delayed base rotation during cornering. In contrast, the active gimbal effectively compensates for these dynamics, ensuring stable tracking across all speed profiles.

\begin{figure}[ht]
\centering
\includegraphics[scale=0.31,page=1]{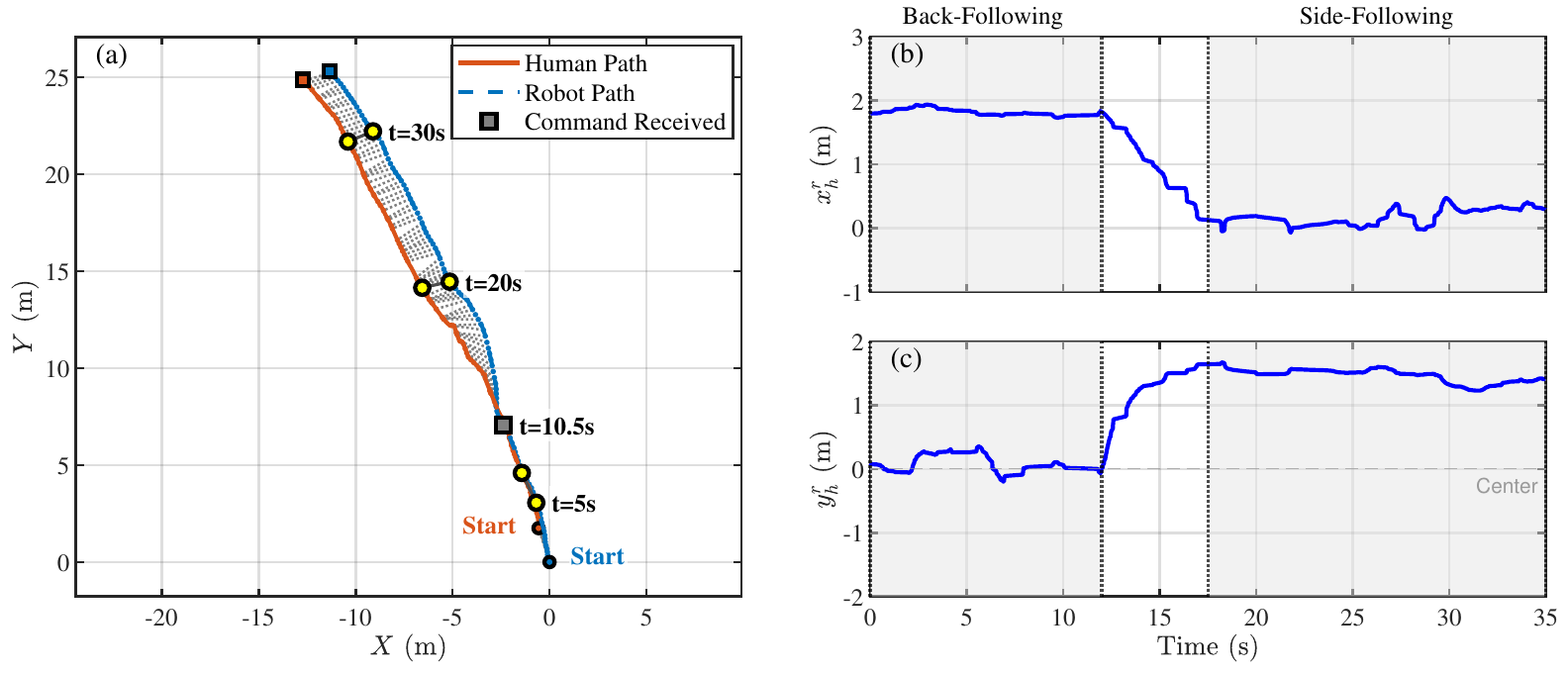}
\caption{Dynamic formation switching from back-following to side-accompaniment: (a) global spatial trajectories, and (b)-(c) relative longitudinal and lateral tracking performance.}
\label{fig:trajectory}
\end{figure}

We validated the LLM interaction layer's ability to dynamically adjust the robot's configuration using implicit natural language. For instance, the utterance, ``I can't see you when you are behind me,'' successfully triggered a transition from back-following to side-by-side accompaniment. Fig.~\ref{fig:trajectory}(a) visualizes the synchronized global trajectories at $0.3\,\mathrm{s}$ intervals. Quantitative profiles in Fig.~\ref{fig:trajectory}(b, c) confirm a stable maneuver: the longitudinal distance ($x_h^r$) converges from $1.8\,\mathrm{m}$ to zero, while the lateral distance ($y_h^r$) smoothly shifts from $0\,\mathrm{m}$ to a $1.5\,\mathrm{m}$ offset.
\begin{table*}[t]
\centering
\caption{End-to-end dialogue assistance results in a simulated elderly-care facility.}
\label{tab:semantic_task}
\renewcommand{\arraystretch}{1.5} 
\resizebox{\textwidth}{!}{%

\begin{tabular}{llllcc}
\toprule
\midrule
\textbf{Utterance Type} & \textbf{Instruction} & \textbf{Target Action} & \textbf{Goal / Context} & \textbf{Rule-based\cite{gupta2025speech}} & \textbf{Proposed} \\ \hline

\multirow{3}{*}{\textbf{Explicit}} &
``Take me to the restroom.'' &
\texttt{A\_navigate} (guide to destination) &
Bathroom &
10/10 & 10/10 \\
&
``Take me to exercise.'' &
\texttt{A\_navigate} (guide to destination) &
Rehabilitation Room &
10/10 & 9/10 \\
&
``Can you stay with me for a while?'' &
\texttt{A\_follow} (active accompaniment) &
-- &
10/10 & 10/10 \\ \hline

\multirow{3}{*}{\textbf{Indirect}} &
``I feel dizzy.'' &
\texttt{A\_alert} (notify nurse) &
-- &
0/10 & 9/10 \\
&
``I'm thirsty.'' &
\texttt{A\_retrieve} (fetch item) &
Resident Room &
0/10 & 9/10 \\
&
``My leg hurts.'' &
\texttt{A\_alert} (notify nurse) &
-- &
0/10 & 10/10 \\ \hline

\multirow{6}{*}{\textbf{Multi-turn}} &
\begin{tabular}{@{}l@{}}
\textbf{U1:} ``Take me to the restroom.'' \\
\textbf{U2:} ``Actually, go to the nurse station instead.''
\end{tabular} &
\begin{tabular}{@{}l@{}}
U1: \texttt{A\_navigate} \\
U2: \texttt{A\_navigate} (revise target)
\end{tabular} &
\begin{tabular}{@{}l@{}}
Bathroom \\
Nurse Station
\end{tabular} &
8/10 & 9/10 \\ \cline{2-6}

&
\begin{tabular}{@{}l@{}}
\textbf{U1:} ``I feel dizzy.'' \\
\textbf{U2:} ``And please fetch my heart medicine.''
\end{tabular} &
\begin{tabular}{@{}l@{}}
U1: \texttt{A\_alert} \\
U2: \texttt{A\_retrieve} (append task)
\end{tabular} &
\begin{tabular}{@{}l@{}}
-- \\
Medication Room
\end{tabular} &
0/10 & 8/10 \\ \cline{2-6}

&
\begin{tabular}{@{}l@{}}
\textbf{U1:} ``Take me to exercise.'' \\
\textbf{U2:} ``Wait, I'm too tired. Just stop here.''
\end{tabular} &
\begin{tabular}{@{}l@{}}
U1: \texttt{A\_navigate} \\
U2: \texttt{A\_stop} (preempt action)
\end{tabular} &
\begin{tabular}{@{}l@{}}
Rehabilitation Room \\
Current Location
\end{tabular} &
6/10 & 10/10 \\
\midrule
\bottomrule
\end{tabular}%
}
\end{table*}
\begin{figure}[t]
\centering
\includegraphics[width=0.78\linewidth]{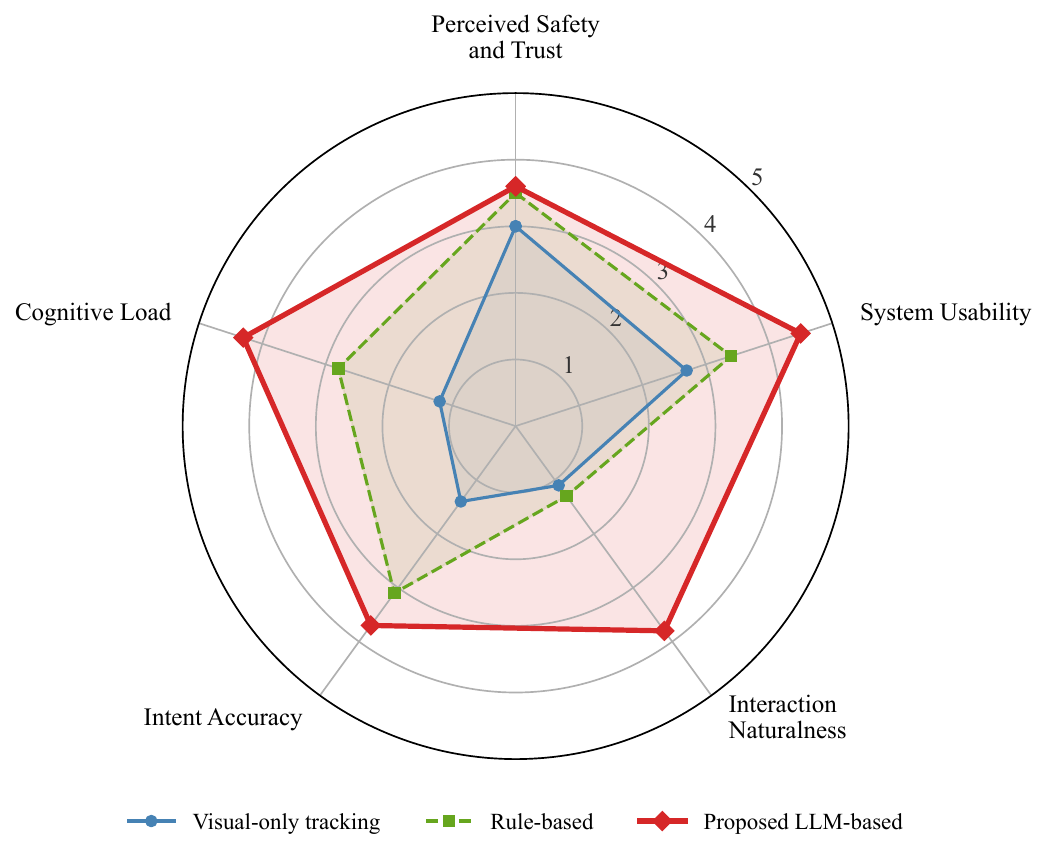}
\caption{User-study results comparing visual-only, rule-based, and the proposed LLM-based methods across five dimensions.}
\label{fig:radar_chart}
\end{figure}
\subsection{Evaluation of Semantic Interaction Tasks in Elderly Care}\label{subsec:exp_semantic}
To evaluate dialogue-driven assistance, we conducted 90 spoken interaction trials across nine instructions (10 repetitions each), covering explicit commands, indirect requests, and multi-turn revisions (Table~\ref{tab:semantic_task}). We compared our LLM-based module against a deterministic, keyword-based baseline~\cite{gupta2025speech}.

Success strictly required correct action selection, valid tool invocation, and successful physical execution, including proper handling of multi-turn context. Overall, our method achieved a 93.3\% (84/90) success rate, significantly outperforming the baseline's 48.9\% (44/90).

While both approaches excelled on explicit commands, the LLM demonstrated clear advantages in semantically underspecified scenarios. For indirect requests (e.g., ``I'm thirsty''), the LLM successfully inferred the intent and invoked $A_{\mathrm{retrieve}}$, whereas the rigid baseline failed entirely. In multi-turn interactions, the LLM maintained dialogue context to dynamically update task plans, achieving a 90\% success rate versus the baseline's 46.7\%. Failure analysis indicated that the few remaining LLM errors resulted from ambiguous phrasing yielding plausible but protocol-incompatible action mappings.

\subsection{User Study}\label{subsec:user_study}
To evaluate interaction quality, ten participants (aged 20–30) tested three human-following modes: visual-only tracking, rule-based interaction, and the proposed LLM-based system. Participants rated the system on a 5-point Likert scale across five dimensions:
\begin{itemize}
    \item Perceived Safety and Trust: Confidence in the robot's safe movement.
    \item System Usability: Ease of operation and interaction.
    \item Interaction Naturalness: Resemblance to human-human communication.
    \item Intent Accuracy: Correct interpretation of user goals.
    \item Cognitive Load: Mental effort required (higher scores denote lower burden).
\end{itemize}

As illustrated in Fig.~\ref{fig:radar_chart}, the proposed LLM-based mode achieved the highest ratings across all dimensions. It excelled in interaction naturalness and intent accuracy by effectively interpreting indirect requests. By eliminating the need for memorized commands, the LLM substantially reduced cognitive demand and offered an intuitive experience. Conversely, the visual-only baseline scored lowest due to its lack of semantic engagement. The rule-based approach provided moderate improvements but was hindered by rigid command templates. Overall, the LLM-driven architecture delivered the most flexible, supportive, and context-aware companion behavior.
\subsection{Evaluation of VLM-Based Fall Detection}\label{subsec:exp_fall}

\begin{figure}[t]
    \centering
    \includegraphics[width=0.95\linewidth]{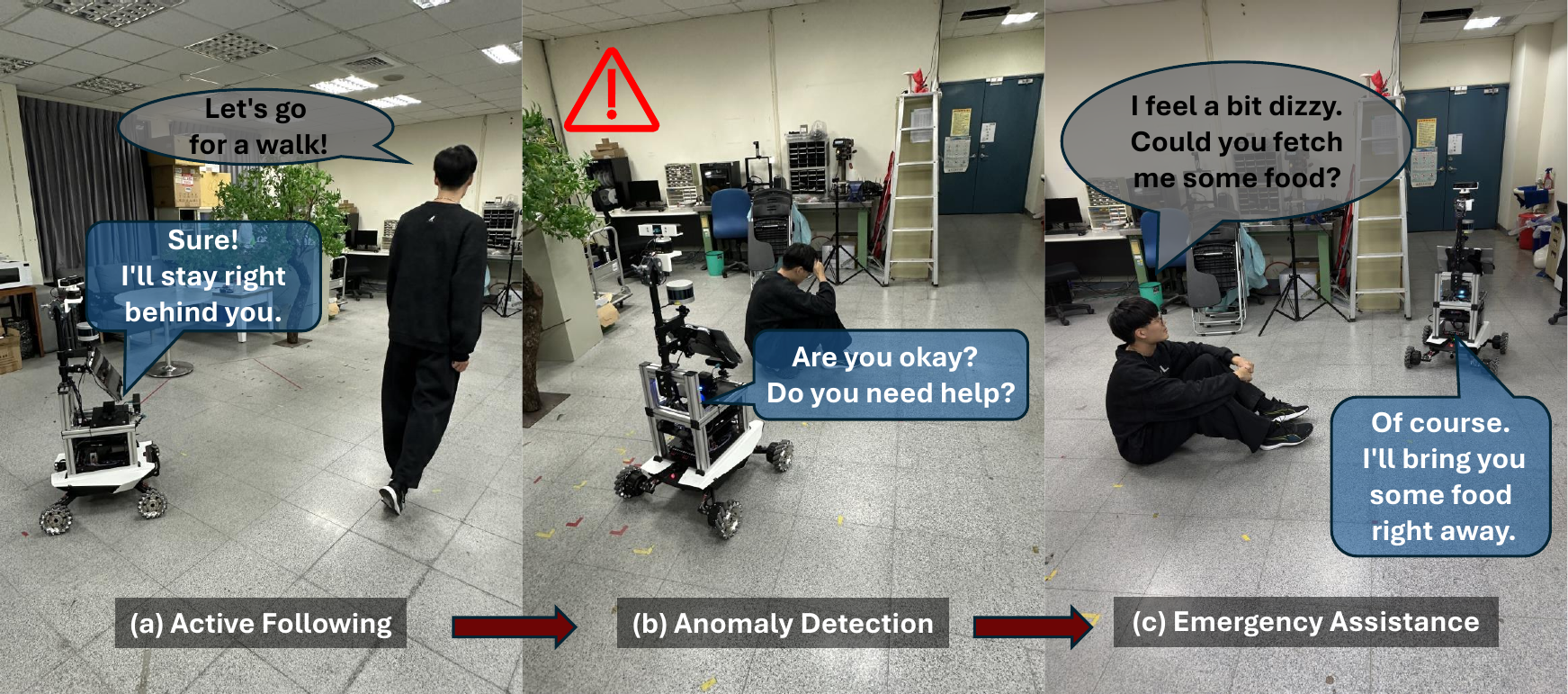}
    \caption{System behavioral transitions during a caregiving scenario, illustrating (a) dynamic human-following, (b) VLM-triggered anomaly detection, and (c) context-aware emergency assistance via LLM-driven dialogue.}
    \label{fig:anomaly detection and emergency response}
\end{figure}

At the system architecture level, the decoupled safety agent utilizes a VLM to continuously monitor the user's postural state via RGB observations during active following. As illustrated in Fig.~\ref{fig:anomaly detection and emergency response}, upon detecting an anomalous posture or fall, the agent immediately preempts the ongoing escorting task and halts the robot's motion to mitigate the risk of secondary injury. Following the physical halt, the LLM-driven interaction layer ensures a seamless transition into emergency assistance mode. It automatically initiates a verbal assessment to ascertain the user's condition and remains fully capable of interpreting post-incident requests. For instance, the system can dynamically schedule a retrieval action ($A_{\mathrm{retrieve}}$) to fetch items upon the user's request, thereby facilitating a context-aware and closed-loop caregiving response.
\section{Conclusion}\label{sec:conclusion}
This paper presented an elderly-care-oriented companion robot that integrates active visual human-following, LLM-driven speech interaction, and VLM-based safety monitoring within a unified ROS~2 framework. The system couples a stereo-vision following module with a PID-controlled gimbal for robust target centering, a GPT Realtime API interaction layer for speech-to-speech dialogue and constrained tool invocation, and a decoupled safety agent capable of preempting ongoing behaviors upon detecting fall-related states. Experimental results demonstrate the effectiveness of the proposed approach in dynamic caregiving scenarios. Future work includes transitioning the system's computational load to local edge devices to minimize latency and reduce dependency on cloud services. Additionally, a real-world deployment and evaluation at an actual elderly care center will also be conducted.
\bibliographystyle{IEEEtran}
\bibliography{reference.bib}

\end{document}